\documentclass[a4paper]{llncs}

\usepackage{amssymb}
\usepackage{amsmath}
\usepackage{graphicx}
\usepackage{hyperref}

\usepackage{array}
\usepackage{multirow}
\usepackage[table]{xcolor}
\definecolor{MyGreen}{rgb}{0,0.91,0.04001}

\newcommand{\keywords}[1]{\par\addvspace\baselineskip
\noindent\keywordname\enspace\ignorespaces#1}

\begin{document}

\mainmatter  

\title{Automatically Segmenting the Left Atrium from Cardiac Images Using Successive 3D U-Nets and a Contour Loss}

\titlerunning{Automatically Segmenting the Left Atrium}
\author{Shuman Jia\inst{1}, Antoine Despinasse\inst{1}, Zihao Wang\inst{1}, Herv\'{e} Delingette\inst{1}, Xavier Pennec\inst{1}, Pierre Ja\"{i}s\inst{2}, Hubert Cochet\inst{2}, \and Maxime Sermesant\inst{1}}
\institute{Universit\'{e} C\^{o}te d'Azur, Epione Research Group, Inria, Sophia Antipolis, France\and IHU Liryc, University of Bordeaux, Pessac, France}

\maketitle

\begin{abstract}
Radiological imaging offers effective measurement of anatomy, which is useful in disease diagnosis and assessment. Previous study \cite{mcgann2014atrial} has shown that the left atrial wall remodeling can provide information to predict treatment outcome in atrial fibrillation. Nevertheless, the segmentation of the left atrial structures from medical images is still very time-consuming. 
Current advances in neural network may help creating automatic segmentation models that reduce the workload for clinicians. 
In this preliminary study, we propose automated, two-stage, three-dimensional U-Nets with convolutional neural network, for the challenging task of left atrial segmentation. 
Unlike previous two-dimensional image segmentation methods, we use 3D U-Nets to obtain the heart cavity directly in 3D. The dual 3D U-Net structure consists of, a first U-Net to coarsely segment and locate the left atrium, and a second U-Net to accurately segment the left atrium under higher resolution. In addition, we introduce a Contour loss based on additional distance information to adjust the final segmentation.
We randomly split the data into training datasets (80 subjects) and validation datasets (20 subjects) to train multiple models, with different augmentation setting. Experiments show that the average Dice coefficients for validation datasets are around 0.91 - 0.92, the sensitivity around 0.90-0.94 and the specificity 0.99. Compared with traditional Dice loss, models trained with Contour loss in general offer smaller Hausdorff distance with similar Dice coefficient, and have less connected components in predictions. Finally, we integrate several trained models in an ensemble prediction to segment testing datasets.

\end{abstract}

\section{Introduction}

Atrial fibrillation (AF) is the most frequently encountered  arrhythmia in  clinical  practice, especially in aged population \cite{zoni2014epidemiology,morillo2017atrial}. It is characterized by uncoordinated electrical activation and disorganized contraction of the atria. This condition is associated with life-threatening consequences, such as heart failure, stroke and vascular cerebral accident. AF also leads to increased public resource utilization and expense on health care.

With evolving imaging technologies, the analysis of cardiovascular diseases and computer-aided interventions has been developing rapidly. Imaging of the heart is routinely performed in some hospital centers when managing AF and prior to atrial ablation therapy, an invasive treatment to establish trans-mural lesions and block the propagation of arrhythmia. Automated segmentation from cardiac images will benefit the studies of left atrial anatomy, tissues and structures, and provide tools for AF patient management and ablation guidance. 

In recent years, with the continuous development of deep learning, neural network models have shown significant advantages in different visual and image processing problems \cite{Cecco2017}. Automatic segmentation of 3D volumes from medical images by deep neural network also attracts increasing attention in the research community of medical image analysis \cite{Litjens2017,Liu2018}.

In this study, we utilize 3D U-Nets with convolutional neural network (CNN), which shows clear advantages compared with traditional feature extraction algorithms \cite{Zeiler2014,Bengio2012}. Based on that, Ronneberger et al. proposed the original U-Net structure \cite{Ronneberger2015} with CNN. Traditional 2D U-Net has achieved good results in the field of medical image segmentation \cite{Ronneberger2015,Li2017}. However, it performs convolution on the 2D slices of the images and cannot capture the spatial relationship between slices.
Its 3D extension \cite{Cicek2016}, expands the filter operator into 3D space. This extracts image features in 3D, and hence takes into account the spatial continuity between slices in medical imaging. This may better reflect shape features of the corresponding anatomy, enabling full use of the spatial information in 3D images.

Previously, Tran et al. used 3D CNNs to extract temporal and spatial features \cite{Tran2015}. They experimented with different sets of data. Hou et al. used 3D CNN to detect and segment pedestrians in a video sequence \cite{Hou2017}. The previous studies show that 3D CNN outperformed 2D CNN when dealing with sequences issues.

3D U-Net was used in \cite{Cicek2016} to realize semi-automatic segmentation of volumetric images. Oktay et al. segmented the ventricle from magnetic resonance (MR) images with 3D U-Net. They introduced an anatomical regularization factor into the model  \cite{Oktay2017}, while we choose to use loss function at pixel level.

In the following sections, we will present the two-stage network to segment the left atrium from MR images. The network consists of two successive 3D U-Nets. The first U-Net is used to locate the segmentation target. The second U-Net performs detailed segmentation from cropped region of interest. We introduce a new loss function, Contour loss for the second U-Net. Results will be shown in Section~\ref{sec:evaluation}.

\section{Method}

\subsection{Dual 3D U-Nets - cropping and segmenting}

\begin{figure}
\centering
\includegraphics[width=\textwidth]{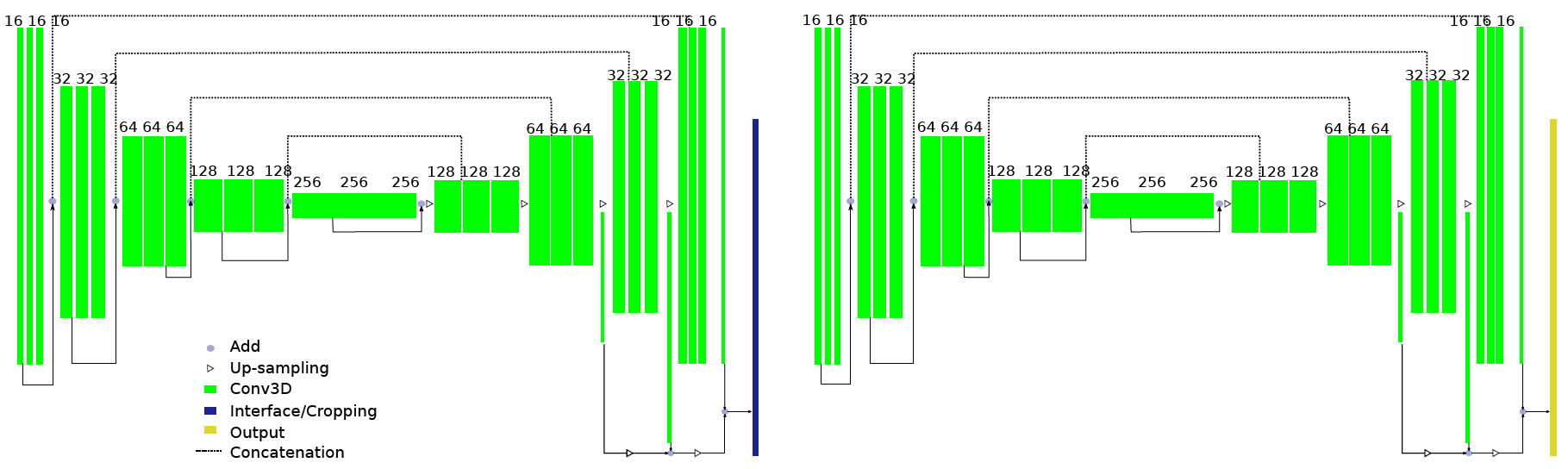}
\caption{Proposed Dual 3D U-Net Structure. Green blocks represent 3D features; Dark blue refers to interface operation to crop the region of interest based on first U-Net prediction.}
\label{fig:my_label}
\end{figure}

U-Net is a typical encoder-decoder neural network structure. The images are encoded by the CNN layers in the encoder. The output characteristics and the feature maps at different feature levels of the encoder serve as input of the decoder. The decoder is an inverse layer-by-layer decoding process. Such a codec structure can effectively extract image features of different levels so as to analyze the images in each dimension.
The 3D U-Net used in this paper is a specialization of 3D U-Net proposed by {\c{C}}i{\c{c}}ek et al \cite{Cicek2016}. The implementation of U-Net follows the work of Isensee et al \cite{isensee2017brain}. We propose a successive dual 3D U-Net architecture, illustrated in Fig.~\ref{fig:my_label}.

The first 3D U-Net locates and coarsely extracts the region of interest. Its input is MR images normalized and re-sized to $[128,128,128]$. Its output is preliminary predicted masks of the left atrium. We keep the largest connected component in the masks, and compute the spatial location of the left atrium. Then, we crop the MR images and ground truth masks with a cuboid centered at the left atrium. 

The second network performs a secondary processing of the cropped images using the full resolution. Because the higher is the resolution, the larger is the needed memory, we keep only the region around the left atrium, so as to preserve information that is essential for left atrial segmentation. But also, this allows to put a higher resolution on the region of interest with the same amount of memory resource. The input for the second U-Net is MR images cropped around the predicted left atrium without re-sampling of size $[224,144,96]$. Its output is our prediction for the left atrial segmentation. We train the second U-Net with two kinds of ground truths, binary segmentation masks $M$ and euclidean distance maps $D(M)$, as shown in Fig~\ref{fig:segmentation1}. Here, we introduce a new loss function based on Contour distance. 

\subsection{Loss Functions}

Using Dice coefficient as loss function can reach high accuracy. 
However, experiments show that, because the inside of the left atrial body accounts for most pixels, the network would stop to optimize when it finds satisfying segmentation of the left atrial body. Instead of the volume inside, the contour is what we want to obtain accurately for the segmentation. It is challenging to segment accurately especially the region around the pulmonary veins and the appendage. Hence, we introduce the Contour distance into the loss function. 

The distance maps $D(M)$ of the ground truth segmentation $M$ illustrate how far is each pixel from the contour of the left atrium. We compute the distance for pixels inside and outside the left atrium, based on the euclidean distance transform algorithm implemented in scipy. The definition of a Hausdorff distance is symmetric between two point sets. But to make it easy to be implemented in neural networks, we do not compute the distance map of the changing prediction $P$ in the training process, and use an unitary distance:
\begin{equation}
loss_{contour} = \sum_{p\in contour(P)}\min_{m\in M}\|p-m\|_2 = \sum (D(M) \circ contour(P)),
\end{equation}
where $\circ$ performs element-wise multiplication. $P$ is the prediction of the U-Net, after sigmoid activation of the output layer. To compute a contour of predicted left atrium, we can apply 3D Sobel filters on $P$ and add the absolute value of results in three directions:
\begin{equation}
contour(P) = \lvert P * s_x \rvert + \lvert P * s_y \rvert + \lvert P * s_z \rvert,
\end{equation}
where $*$ denotes 3D convolution operation and $+$ denotes pixel-wise addition. $s_x$, $s_y$ and $s_z$ are 3D Sobel–Feldman kernels in x-, y- and z-direction, for example: 
\begin{align*}
s_z =  \left[
 \begin{bmatrix}
  +1 & +2 & +1 \\
  +2 & +4 & +2 \\
  +1 & +2 & +1
 \end{bmatrix} , 
  \begin{bmatrix}
  0  & \ 0 & \ 0 \\
  0  & \ 0 & \ 0 \\
  0  & \ 0 & \ 0
 \end{bmatrix}, 
  \begin{bmatrix}
  -1 & -2 & -1 \\
  -2 & -4 & -2 \\
  -1 & -2 & -1
 \end{bmatrix}
 \right].
\end{align*}

In the optimization process, the Contour loss decreases when the contour of prediction is nearer that of the ground truth. However, if $D(M)$ and $contour(P)$ are both always positive, a bad global minimal exists: let prediction $P$ remain constant so that $contour(P)\sim0$. To avoid this, we add a $drain$ on distance maps. For example, we set $drain = -1.8$, and therefore $D(M) + drain$ has negative value on the contour of $M$, since all pixels on the contour of $M$ have a euclidean distance $\leqslant\sqrt 3$. This creates a drain channel for the model to continue the optimization process towards negative loss value.
\begin{equation}
loss_{contour} = \sum ((D(M)+drain) \circ contour(P)).
\end{equation}
The loss function is differentiable and converging in 3D U-Net training trials.\footnote{The code is available on \url{https://gitlab.inria.fr/sjia/unet3d_contour}.}

\subsection{Ensemble Prediction}
Contour loss provides a spatial distance information for the segmentation, while Dice coefficient measures the volumes inside the contour. We combine the two loss functions in a ensemble prediction model.

\begin{figure}
\centering
\includegraphics[width = \textwidth]{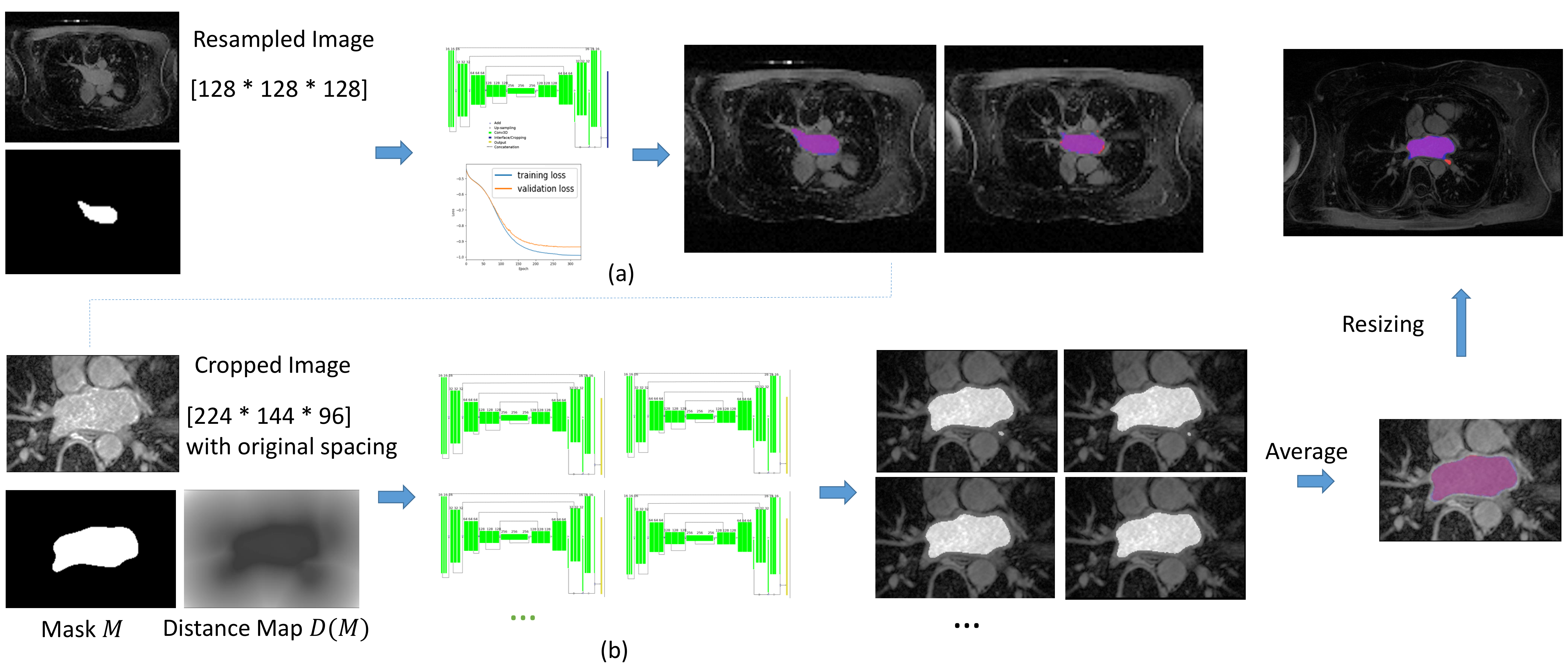}
\caption{The framework of successive U-Nets training. (a) The fist U-Net - cropping; (b) the second U-Net - segmenting, with ensemble prediction models. We show here axial slices of MR images, overlapped with manual segmentation of the left atrium in blue, our segmentation in red, intersection of the two in purple.}
\label{fig:segmentation1}
\end{figure}

We visualize the process in Fig.~\ref{fig:segmentation1}.

\subsubsection{Fixed Experimental Setting}

In this study, we set batch size to $1$. Training-validation split ratio equals to 0.8. We perform normalization of image intensity. The initial learning rate of our neural network is $5e^{-4}$. Learning rate will be reduced to half after 10 epochs if the validation loss is not improving. The early convergence is defined as no improvement after 50 epochs. Number of base filters is 16. Maximum number of epochs is 500, and 200 steps per epoch. Using large initial learning rate reduces the time to find the minimum and may also overstep the local minima to converge closer to the global minimum of the loss function. However, the accuracy of segmentation also relies on when we stop the training to avoid over-fitting.

We use computation cluster with GPU capacity 6.1 for training. The first U-Net takes around 200 - 300 epochs to reach convergence. We extract the largest connected components in predictions of the first U-Net to crop the original MRI images around the left atrium, without re-sampling, of size $[224, 144, 96]$. The second U-Net takes around 150 - 200 epochs to reach convergence. Then the prediction results of the second U-Net are re-sized without re-sampling to the original image size.

\subsubsection{Varying Experimental Setting}\label{sec:setting}
For the second U-Net, we change on the one hand, the options for augmentation: horizontal/vertical flip set to True or False; rotation range set to 0, 7, 10 degree; width/height shift range set to 0.0 - 0.1; zoom in or not, zoom range set to 0.1, 0.2. On the other hand, we alter the loss function option, Dice coefficient and Contour loss. We choose multiple trained U-Net models with above experimental settings for ensemble prediction. We also train twice with some parameters but with different validation splitting. We make the final decision of segmentation based on the average of all predictions, similar to letting multiple agents vote for each pixel if it belongs to left atrium or not, in majority voting system.

\section{Evaluation on Clinical Data}\label{sec:evaluation}

\subsection{Materials}
A total of $100$ 3D GE-MRIs from patients with AF are provided by the STACOM 2018 Atrial Segmentation Challenge. The original resolution of the data is $0.625 \times 0.625 \times 0.625\ mm^3$. 3D MR images were acquired using a clinical whole-body MRI scanner and the corresponding ground truths of the left atrial masks were manually segmented by experts in the field. 

\subsection{Comparison of Loss Functions}
We assess the segmentation results of individual models, trained with different experimental setting, as described in Sec.~\ref{sec:setting}, to compare the prediction performance using Dice loss and Contour loss.

\subsubsection{Evaluation Metrics}
The evaluation metrics are Dice coefficient, Hausdorff distance (HD) and confusion matrix, as shown in Table~\ref{table:confusion}. Multiple evaluation metrics provide us different views to assess the performance of models.
\begin{table}
\setlength{\tabcolsep}{2.6pt}
\renewcommand{\arraystretch}{1.2}
\centering
\caption{Evaluation of validation datasets segmentations using two loss functions: Dice coefficient loss (top) and Contour loss (bottom).}\label{table:confusion}
\begin{tabular}{|*{7}{c|}}\hline 
    \multirow{2}{*}{} & \multirow{3}{*}{} & \multicolumn{2}{c|}{Confusion Matrix}  & Dice  & \multicolumn{2}{c|}{Contour distance (pixel)} \\
    &   & Sensitivity  & Specificity   &   & Average HD   & HD \\ 
    \hline 
    \multirow{8}{*}{\shortstack{Model\\(Dice loss)}}   
    & 1 & \cellcolor{MyGreen!65}$0.93 \pm 0.03$ & \cellcolor{MyGreen!15}$0.99 \pm 0.00$ & \cellcolor{MyGreen!10}$0.910 \pm 0.032$ & \cellcolor{MyGreen!5}$1.71 \pm 0.53$ & \cellcolor{MyGreen!5}$34.29 \pm 15.00$ \\ 
    & 2 & \cellcolor{MyGreen!80}$0.94 \pm 0.04$ & \cellcolor{MyGreen!15}$0.99 \pm 0.00$ & \cellcolor{MyGreen!25}$0.913 \pm 0.029$ & \cellcolor{MyGreen!19}$1.63 \pm 0.57$ & \cellcolor{MyGreen!30}$27.39 \pm 12.19$ \\ 
    & 3 & \cellcolor{MyGreen!35}$0.91 \pm 0.05$ & \cellcolor{MyGreen!15}$0.99 \pm 0.00$ & \cellcolor{MyGreen!30}$0.914 \pm 0.023$ & \cellcolor{MyGreen!17}$1.62 \pm 0.42$ & \cellcolor{MyGreen!20}$30.04 \pm 11.76$ \\ 
    & 4 & \cellcolor{MyGreen!50}$0.92 \pm 0.04$ & \cellcolor{MyGreen!15}$0.99 \pm 0.00$ & \cellcolor{MyGreen!55}$0.920 \pm 0.030$ & \cellcolor{MyGreen!56}$1.41 \pm 0.51$ & \cellcolor{MyGreen!52}$23.50 \pm 12.34$ \\ 
    & 5 & \cellcolor{MyGreen!35}$0.91 \pm 0.05$ & \cellcolor{MyGreen!15}$0.99 \pm 0.00$ & \cellcolor{MyGreen!45}$0.918 \pm 0.019$ & \cellcolor{MyGreen!46}$1.47 \pm 0.47$ & \cellcolor{MyGreen!64}$21.66 \pm 7.75$ \\ 
    & 6 & \cellcolor{MyGreen!35}$0.91 \pm 0.04$ & \cellcolor{MyGreen!15}$0.99 \pm 0.00$ & \cellcolor{MyGreen!60}$0.921 \pm 0.024$ & \cellcolor{MyGreen!47}$1.45 \pm 0.48$ & \cellcolor{MyGreen!35}$26.52 \pm 16.14$ \\ 
    & 7 & \cellcolor{MyGreen!50}$0.92 \pm 0.03$ & \cellcolor{MyGreen!15}$0.99 \pm 0.00$ & \cellcolor{MyGreen!68}$0.923 \pm 0.027$ & \cellcolor{MyGreen!58}$1.40 \pm 0.67$ & \cellcolor{MyGreen!52}$23.38 \pm 12.03$ \\ 
    & 8 & \cellcolor{MyGreen!65}$0.93 \pm 0.04$ & \cellcolor{MyGreen!15}$0.99 \pm 0.00$ & \cellcolor{MyGreen!78}$0.927 \pm 0.024$ & \cellcolor{MyGreen!78}$1.28 \pm 0.46$ & \cellcolor{MyGreen!58}$22.48 \pm 16.35$ \\ 
    \hline
    \multirow{8}{*}{\shortstack{Model\\(Contour loss)}}   
    & 1 & \cellcolor{MyGreen!20}$0.90 \pm 0.05$ & \cellcolor{MyGreen!15}$0.99 \pm 0.00$ & \cellcolor{MyGreen!15}$0.911 \pm 0.020$ & \cellcolor{MyGreen!36}$1.60 \pm 0.41$ & \cellcolor{MyGreen!79}$19.80 \pm 6.82$ \\ 
    & 2 & \cellcolor{MyGreen!50}$0.92 \pm 0.04$ & \cellcolor{MyGreen!15}$0.99 \pm 0.00$ & \cellcolor{MyGreen!5}$0.906 \pm 0.045$ & \cellcolor{MyGreen!36}$1.60 \pm 0.58$ & \cellcolor{MyGreen!85}$18.88 \pm 7.13$ \\ 
    & 3 & \cellcolor{MyGreen!35}$0.91 \pm 0.04$ & \cellcolor{MyGreen!15}$0.99 \pm 0.00$ & \cellcolor{MyGreen!40}$0.917 \pm 0.025$ & \cellcolor{MyGreen!44}$1.48 \pm 0.51$ & \cellcolor{MyGreen!65}$21.57 \pm 8.13$ \\ 
    & 4 & \cellcolor{MyGreen!50}$0.92 \pm 0.04$ & \cellcolor{MyGreen!40}$1.00 \pm 0.00$ & \cellcolor{MyGreen!40}$0.917 \pm 0.024$ & \cellcolor{MyGreen!58}$1.40 \pm 0.33$ & \cellcolor{MyGreen!73}$20.21 \pm 7.53$ \\ 
    & 5 & \cellcolor{MyGreen!20}$0.90 \pm 0.04$ & \cellcolor{MyGreen!40}$1.00 \pm 0.00$ & \cellcolor{MyGreen!50}$0.919 \pm 0.024$ & \cellcolor{MyGreen!36}$1.53 \pm 0.51$ & \cellcolor{MyGreen!42}$24.81 \pm 6.57$ \\
    & 6 & \cellcolor{MyGreen!5}$0.89 \pm 0.03$ & \cellcolor{MyGreen!40}$1.00 \pm 0.00$ & \cellcolor{MyGreen!60}$0.921 \pm 0.019$ & \cellcolor{MyGreen!31}$1.58 \pm 0.59$ & \cellcolor{MyGreen!45}$24.04 \pm 11.72$ \\
    & 7 & \cellcolor{MyGreen!35}$0.91 \pm 0.04$ & \cellcolor{MyGreen!15}$0.99 \pm 0.00$ & \cellcolor{MyGreen!30}$0.914 \pm 0.021$ & \cellcolor{MyGreen!34}$1.56 \pm 0.47$ & \cellcolor{MyGreen!60}$22.65 \pm 7.52$ \\
    & 8 & \cellcolor{MyGreen!50}$0.92 \pm 0.04$ & \cellcolor{MyGreen!40}$1.00 \pm 0.00$ & \cellcolor{MyGreen!40}$0.917 \pm 0.030$ & \cellcolor{MyGreen!56}$1.41 \pm 0.45$ & \cellcolor{MyGreen!80}$19.66 \pm 4.52$ \\
    \hline
\end{tabular}
\end{table}

The Dice index for predicted segmentation in validation datasets attained 0.91 - 0.92. The sensitivity of prediction was around 0.90 - 0.94 and the specificity 0.99. The proposed method closely segmented the atrial body, with both loss functions, compared with manual segmentation. Different from traditional Dice loss, models trained with Contour loss in general offered smaller Hausdorff distance with similar Dice coefficient.

\subsubsection{Visualization}
\begin{figure}
\centering
\includegraphics[width = \textwidth]{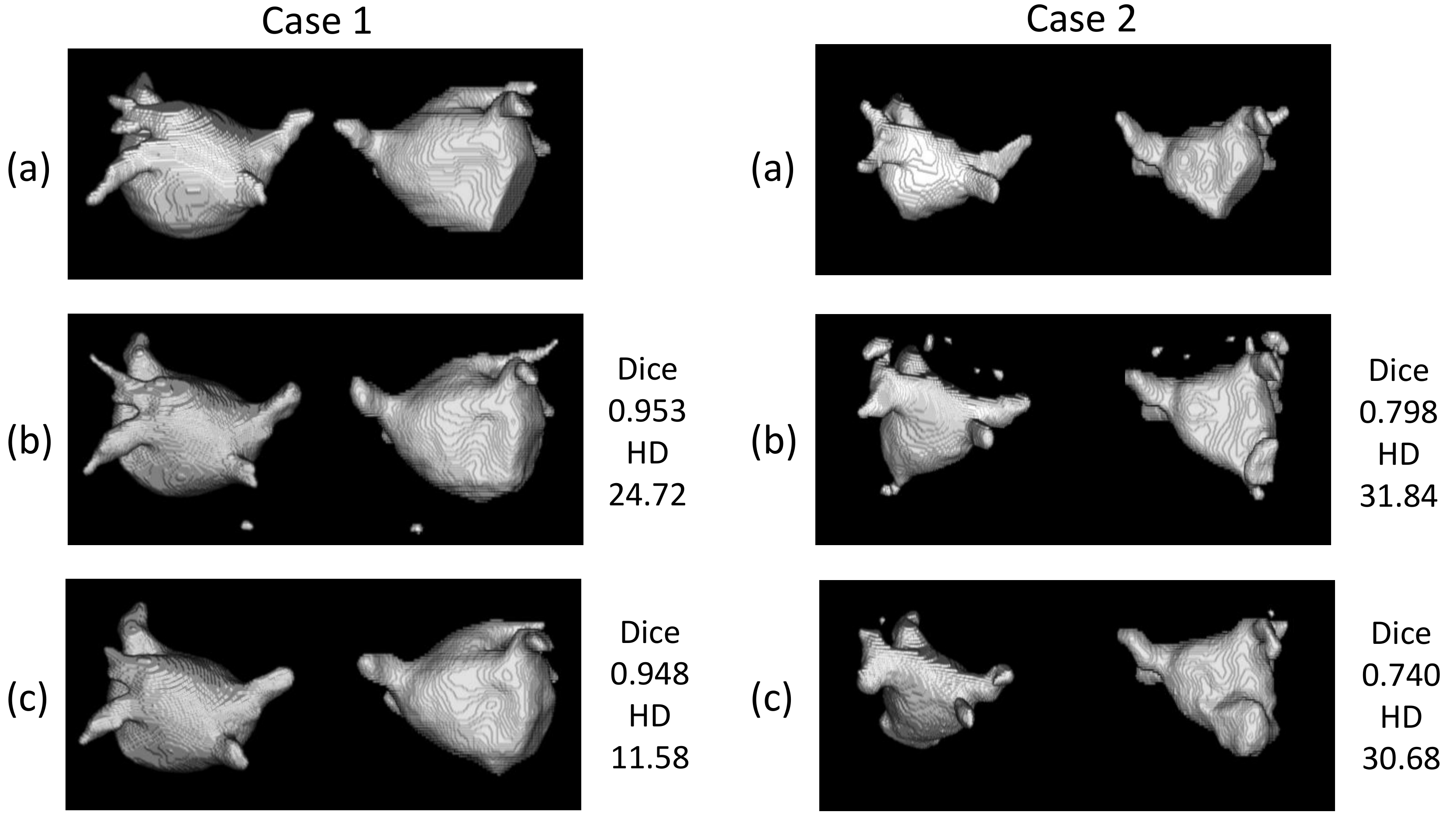}
\caption{Visualization of good and bad validation datasets segmentations, with their Dice, Hausdorff distance (HD) with respect to the ground truth. (a) manually segmented; (b) predicted with Dice coefficient loss; (c) predicted with Contour loss.}
\label{fig:segmentation2}
\end{figure}

We visualize the predicted segmentation results of validation datasets in Fig.~\ref{fig:segmentation2} in a 3D view. Case 1 and Case 2 are selected to represent respectively, good scenario and bad scenario.
For the two loss functions, differences lay in the boundary, the region close to the pulmonary veins and septum. With Dice loss function, there were more details and sharp edges, and therefore more disconnected spots not belonging to the left atrium. With the Contour loss function, the smoothness of the contour, and shape consistency were better maintained.

\subsubsection{Connected Components}
The left atrium should be a single connected component in binary segmentation mask. We present in Table~\ref{table:components} the number of connected components in raw predictions given by U-Nets and compare the two loss functions. Using Dice coefficient loss alone produced more disconnect components not belonging to the left atrial structures.

\begin{table}
\setlength{\tabcolsep}{2.6pt}
\renewcommand{\arraystretch}{1.2}
\centering
\caption{Number of connected components in predicted segmentations.}\label{table:components}
\begin{tabular}{|*{3}{c|}}\hline 
    & Mean $\pm\ \sigma$ & Maximum \\
    \hline
    Model (Dice loss) & $2.85 \pm 2.83$  &  20\\
    \hline
    Model (Contour loss) & $1.28 \pm 0.65$  & 5 \\\hline 
\end{tabular}
\end{table}

\subsubsection{Ensemble Prediction}
To reduce the irregular bumps and disconnected components, we choose in total 11 trained U-Net models with both loss functions, to perform an ensemble prediction for testing datasets. We add the probabilistic segmentations of all U-Nets and threshold their sum $\geq 5.5$.

With Dice loss, more details can be captured. With Contour loss, the shapes look more regular and smoother globally. The difficult regions to segment remain the septum and especially the appendage for both loss functions. The manual segmentations are usually performed slice-by-slice, there exists sudden discontinuity between two axial slices. While our segmentation is based on 3D operators, the segmented region is continuous between slices, which accounts for part of mismatch between manual segmentation and network segmentation that cannot be avoided. 

\section{Conclusion}

In this paper, we proposed a deep neural network with dual 3D  U-Net structure, to segment the left atrium from MR images. To take into consideration the shape characteristics of the left atrium, we proposed to include distance information and created a Contour loss function. Using multiple trained models in an ensemble prediction can improve the performance, reducing the impact of accidental factors in neural network training. 

Based on previous studies on cardiac cavities segmentation, our method accurately located the region of interest and provided good segmentations of the left atrium. Experiments show that the proposed method well captured the anatomy of the atrial volume in 3D space from MR images. The new loss function achieved a fine-tuning of contour distance and good shape consistency.

The automated segmentation model can in return reduce manual work load for clinicians and has promising applications in clinical practice. Potential future work includes integrating the segmentation model into a clinic-oriented AF management pipeline.

\subsubsection*{Acknowledgments.} Part of the research was funded by the Agence Nationale de la Recherche (ANR)/ERA CoSysMed SysAFib project. This work was supported by the grant AAP Santé 06 2017-260 DGA-DSH, and by the Inria Sophia Antipolis - Méditerranée, NEF computation cluster. The author would like thank the work of relevant engineers and scholars.

\end{document}